%% file: 315.tex
\documentclass[10pt,twocolumn,letterpaper]{article}

\usepackage{wacv}
\usepackage{times}
\usepackage{epsfig}
\usepackage{graphicx}
\usepackage{amsmath}
\usepackage{amssymb}

\usepackage{tablefootnote}
\usepackage{multirow}
\usepackage{float}
\usepackage{epsfig}
\usepackage{graphicx}
\usepackage{amsmath}
\usepackage{amssymb}
\usepackage{graphicx, caption, subcaption}
\usepackage{xcolor}
\usepackage{footnote}
\usepackage{algorithm}
\usepackage{algpseudocode}

\graphicspath{{./figures/}{./figures/draw/}}

\def\u{{\mathbf u}}
\def\f{{\mathbf f}}
\def\bb{{\mathbf b}}

\def\W{{\mathbf W}}
\def\B{{\mathbf B}}
\def\F{{\mathbf F}}
\newcommand\norm[1]{\left\lVert#1\right\rVert}

\def\junk#1{\def\@junk{#1}}

\wacvfinalcopy 

\def\wacvPaperID{315} 

\ifwacvfinal\fi
\begin{document}

\title{Binary Constrained Deep Hashing Network for Image Retrieval\\ without Manual Annotation}

\author{Thanh-Toan Do$^{\dagger}$
\and Tuan Hoang$^{\ddagger}$
\and Dang-Khoa Le Tan$^{\ddagger}$
\and Trung Pham$^{\star}$
\and Huu Le$^{\star}$
\and Ngai-Man Cheung$^{\ddagger}$
\and Ian Reid$^{\star}$\\
\and $^{\dagger}$University of Liverpool,  $^{\ddagger}$Singapore University of Technology and Design, $^{\star}$University of Adelaide \\
{\tt\small thanh-toan.do@liverpool.ac.uk}, \tt\small nguyenanhtuan\_hoang@mymail.sutd.edu.sg, \\ \tt\small \{letandang\_khoa,ngaiman\_cheung\}@sutd.edu.sg,  \tt\small \{trung.pham,huu.le,ian.reid\}@adelaide.edu.au\\
}
\maketitle
\ifwacvfinal\thispagestyle{empty}\fi

\input{tex/abstract.tex}
\input{tex/intro.tex}

\input{tex/relatedwork.tex}

\input{tex/proposition.tex}

\input{tex/experiment.tex}
\input{tex/conclusion2.tex}

{\small
\bibliographystyle{ieee}
\bibliography{sigprocref}
}

\end{document}

%% file: tex/abstract.tex
\begin{abstract}

Learning compact binary codes for image retrieval task using deep neural networks has attracted increasing attention recently. However, training deep hashing networks for the task is challenging due to the binary constraints on the hash codes, the similarity preserving property, and the requirement for a vast amount of labelled images. To the best of our knowledge, none of the existing methods has tackled all of these challenges completely in a unified framework.  In this work, we propose a novel end-to-end deep learning approach for the task, in which the network is trained to produce binary codes directly from image pixels without the need of manual annotation. 
In particular, to deal with the non-smoothness of binary constraints, we propose a novel pairwise constrained loss function, which simultaneously encodes the distances between pairs of hash codes, and the binary quantization error. 
In order to train the network with the proposed loss function, we propose an efficient parameter learning algorithm. 
In addition, to provide similar / dissimilar training images to train the network, we exploit 3D models reconstructed from unlabelled images for automatic generation of enormous training image  pairs.  The extensive experiments on image retrieval benchmark datasets demonstrate the improvements of the proposed method over the state-of-the-art compact representation methods on the image retrieval problem.  

\end{abstract}

%% file: tex/intro.tex
\section{Introduction}
We are interested in learning compact image representations for large scale content-based image retrieval problem. 
Recent researches have applied deep learning to image retrieval problem and achieved  improvements in comparison to traditional local feature approaches. In ~\cite{DBLP:conf/cvpr/RazavianASC14,DBLP:conf/cvpr/AzizpourRSMC15}, the authors show that using the real-valued features from off-the-shelf pretrained networks to represent images achieve impressive retrieval results. In~\cite{babenko2014neural,netvlad,DBLP:journals/ijcv/GordoARL17} the authors further show that fine-tuning pretrained deep networks for image retrieval task helps to boost the retrieval performance. However, to fine-tune a deep network, it requires an enormous amount of labelled images which is  
not easy to achieve. It is because annotating images with labels or tags requires skilled manpower, and the label of an image is not always well defined. In addition, although representing images by real-valued high dimensional features from deep networks achieves high    retrieval accuracy,  it is not applicable for large scale retrieval problem. It is because these representations cause expensive storage and time-consuming searching. 

Using binary hash codes to represent images is an attractive approach for large scale vision problems including image retrieval because the binary codes allow the fast computation and efficient storage~\cite{BA_CVPR15,DBLP:conf/cvpr/GongL11,DBLP:conf/cvpr/ZhaoHWT15,DBLP:conf/cvpr/LaiPLY15,DBLP:conf/cvpr/Liu0SC16,UH-BDNN-Do2016-0,DBLP:conf/cvpr/HeWS13,Kevin_TPAMI17,DeepHash_TIP17,DBLP:journals/tmm/ErcoliBB17,deepbit2016,deepbit2018,DBLP:journals/tip/ZhangLZZZ15}.
However, learning binary codes in deep networks is challenging. This is because one has to deal with the binary constraint on the hash codes. 
Furthermore, another important requirement of hashing is the similarity preservation, i.e., similar/dissimilar images should have similar/dissimilar binary codes. To achieve this requirement
under deep models, previous deep hashing methods require a vast of amount of manually well-defined labelled datasets to supervise the training or fine-tuning. 
 Unfortunately, such large labelled datasets are not always available,
especially in some problems which are not directly based on classification such as the image retrieval. 

In this paper, we aim to address the above challenges by learning an end-to-end deep neural
network to produce binary hash codes directly from images without the need of manually labelled datasets. 
 In particular, we propose a pairwise binary constrained loss function to model the relative similarities between pairs of hash codes and to encourage the network outputs to be binary values. In order to train the network with the proposed loss, we propose a  novel learning scheme that is inspired from the  penalty method and alternating optimization. 
Furthermore, since our loss function is pairwise, it only requires relative relationship for
pairs of images, i.e., matching and non-matching images. Clearly such relationship
can be obtained without resorting semantic labels. 
Inspired by the recent works~\cite{netvlad,MAC2} 
we exploit 3D models built from unlabelled images using Structure-from-Motion (SfM) to automatically create training data. 
As a result, the training of our deep hashing network can be done completely and automatically by using unlabelled images as inputs. 

In summary, we make the following contributions.
1) We propose a novel end-to-end deep learning framework for learning compact
binary codes in which the input for training the framework is only unlabelled
images. 
2) We propose a novel pairwise loss function that simultaneously encodes the distances between hash codes and the binary quantization error. To train the network with the proposed loss, we develop an efficient alternating optimization to optimize the network parameters. 3) We exploit reconstructed 3D models to automatically create the training data. To the best of our knowledge, this work is the first one that relies on 3D geometry to create data for training an end-to-end deep hashing framework.  
4) We perform solid experiments on three image retrieval datasets to
demonstrate the improvements of the proposed method over the state of the art.

The remainder of the paper is organized as follows: Section \ref{sec:relatedwork} discusses related works. Sections \ref{sec:proposed} and \ref{sec:exp} present the proposed method and experimental results, respectively. Section \ref{sec:concl} concludes the paper.

%% file: tex/relatedwork.tex
\section{Related Work}
\label{sec:relatedwork}
In this section, we review previous works related to the context of our work. Those include traditional hashing approaches, deep hashing approaches, and end-to-end image retrieval with weakly supervised / unsupervised fine-tuning. 
\\
\textbf{Traditional hashing methods:}
Existing binary hashing methods can be categorized as data-independent and data-dependent schemes \cite{DBLP:journals/corr/WangLKC15,DBLP:journals/corr/WangSSJ14_journal,Grauman_review}. 
Data-independent hashing methods~\cite{lsh_vldb09,KLSH_iccv09,KLSH_nips09,DBLP:journals/pami/KulisJG09} rely on random projections for constructing hash functions. Although representative data-independent hashing methods such as Locality-Sensitive Hashing (LSH)~\cite{lsh_vldb09} and its kernelized versions~\cite{KLSH_iccv09,KLSH_nips09} have theoretical guarantees that the more similar data would have higher probability to be mapped into similar binary codes, they require long codes to achieve high precision. 
Different from data-independent approaches, data-dependent hashing methods use available training data for learning hash functions in unsupervised or supervised manner and they usually achieve better retrieval results than data-independent methods. The unsupervised hashing methods~\cite{DBLP:conf/nips/WeissTF08,DBLP:conf/cvpr/GongL11,DBLP:conf/cvpr/HeWS13,BA_CVPR15,CVPR12:SphericalHashing} try to preserve the neighbor similarity of samples in Hamming space without semantic label information. The representative unsupervised hashing methods include Iterative Quantization (ITQ)~\cite{DBLP:conf/cvpr/GongL11}, Spherical Hashing (SPH)~\cite{CVPR12:SphericalHashing}, K-means Hashing (KMH)~\cite{DBLP:conf/cvpr/HeWS13}, etc. 
The supervised hashing methods~\cite{Kulis_learningto,CVPR12:Hashing,CVPR2014Lin,Shen_2015_CVPR,DBLP:conf/icml/NorouziF11} try to preserve the label similarity of samples using labelled training data. The representative supervised hashing methods include Kernel-Based Supervised Hashing (KSH)~\cite{CVPR12:Hashing}, Semi-supervised Hashing (SSH)~\cite{DBLP:journals/pami/WangKC12}, Supervised Discrete Hashing (SDH)~\cite{Shen_2015_CVPR}, Binary Reconstructive Embedding (BRE)~\cite{Kulis_learningto} etc. 
\\
\textbf{Deep hashing methods:}
All the previous hashing methods are originally designed and experimented on hand-crafted features which may limit their performance in practical applications. Recently, to leverage the power of deep convolutional neural networks (CNNs)~\cite{Lecun98gradient-basedlearning,DBLP:conf/nips/KrizhevskySH12,Simonyan14c}, few deep unsupervised hashing and many deep supervised hashing methods have been proposed. 

There are few deep models which are proposed for unsupervised hashing~\cite{DBLP:conf/aaai/SongHGXS18,deepbit2016,deepbit2018,SADH}.  In \cite{deepbit2016,deepbit2018}, the authors proposed an end-to-end deep learning  framework for unsupervised hashing. The network is trained to produce hash codes that minimize the quantization loss with the output of the last VGG's~\cite{Simonyan14c} fully connected layer. 
In~\cite{DBLP:conf/aaai/SongHGXS18}, the authors proposed a region-based deep hashing method in which the network consists of three modules, i.e., object proposal generation, feature extraction, and a hashing layer. 
In~\cite{SADH} the authors proposed an unsupervised deep hashing method that alternatingly proceeds over three training modules: deep hash model training, similarity graph construction, and binary code optimization.
 
Different from unsupervised setting, there are many methods have been proposed for deep supervised hashing. In~\cite{DBLP:conf/aaai/XiaPLLY14} the authors proposed a two-step supervised hashing method which learns a deep CNN based hash function with the pre-computed binary codes. 
Recently, many works proposed end-to-end deep supervised hashing methods in which the image features and the hash codes are simultaneously learned~\cite{DBLP:conf/cvpr/ZhaoHWT15,DBLP:conf/cvpr/LaiPLY15,DBLP:journals/tip/ZhangLZZZ15,Kevin_TPAMI17,DBLP:conf/cvpr/Liu0SC16,DBLP:conf/cvpr/ZhuangLSR16}.
 Most of those models consist of a deep CNN for image feature extraction and a hashing layer that tries to approximate the \textit{sign} function. 
By joint optimization, the produced hash codes have been shown more sufficient to preserve the semantic similarity between images. 
Ideally, the hashing layer should adopt a $sign$ activation function to output exactly binary codes. However, due to the vanishing gradient difficulty of the $sign$ function, an approximation procedure must be employed. For example, $sign$ can be approximated by a tanh-like function $y=tanh(\beta x)$, where $\beta$ is a free parameter controlling the trade off between the smoothness and the binary quantization loss \cite{DBLP:journals/tip/ZhangLZZZ15}. However, it is non-trivial to determine an optimal $\beta$. A small $\beta$ causes large binary quantization loss while a big $\beta$ makes the output of the function close to the binary values, but the gradient of the function almost vanishes, making back-propagation infeasible. 
The problem still remains when the sigmoid-like functions \cite{DBLP:conf/cvpr/ZhaoHWT15,DBLP:conf/cvpr/LaiPLY15,Kevin_TPAMI17,DBLP:conf/cvpr/ZhuangLSR16} are used. 
Another drawback of deep supervised hashing networks is the requirement for a large amount of semantic annotated data which will be used for encoding the semantic similarities in the loss function. However, such large annotated data is usually unavailable in the large scale image retrieval problem. 
\\
\textbf{Weakly supervised / unsupervised fine-tuning:}
In the last few years, image retrieval has witnessed an increasing of performance due to better image representations. In particular, 
features obtained from pretrained CNN models,  which are trained on image classification task, are usually adopted. For example,~\cite{DBLP:journals/corr/ToliasSJ15,DBLP:conf/cvpr/AzizpourRSMC15} used convolutional features, while ~\cite{DBLP:conf/cvpr/RazavianASC14} used fully connected features for the image retrieval. Fine-tuning the pretrained networks has also shown further improvements~\cite{babenko2014neural}.
However, the fine-tuning requires the availability of annotated data. 
Unfortunately, it would be difficult and expensive to manually label a large collection of
images. To overcome this challenge, the recent works proposed to use weakly supervised or unsupervised fine-tuning.
In~\cite{netvlad}, to prepare the data to fine-tune the network for the place recognition task, the authors used a weakly supervised approach, in which they used Google Street View Time Machine for getting GPS-tagged panoramic images taken at nearby spatial locations on the map. Two images taken far from each other are considered as non-matching, while the matching images are selected by the most similar nearby images. In \cite{DBLP:journals/corr/RadenovicTC16,MAC2}, the authors made further improvements over \cite{netvlad}, i.e., discovering matching and non-matching pairs in a totally unsupervised manner. Using a large amount of images downloaded from Flickr with keywords of popular landmarks and cities, they 
applied Structure from Motion \cite{DBLP:conf/cvpr/RadenovicSJFCM16}
for building multiple 3D models. 
Images belonging to the same 3D model and sharing enough 3D points are considered as matching, while images belonging to different 3D models are considered as non-matching.

%% file: tex/proposition.tex
\section{Binary Constrained Deep Hashing Network without Manual Annotation}
\begin{figure*}[t]
\centering
\includegraphics[width=0.99\textwidth]{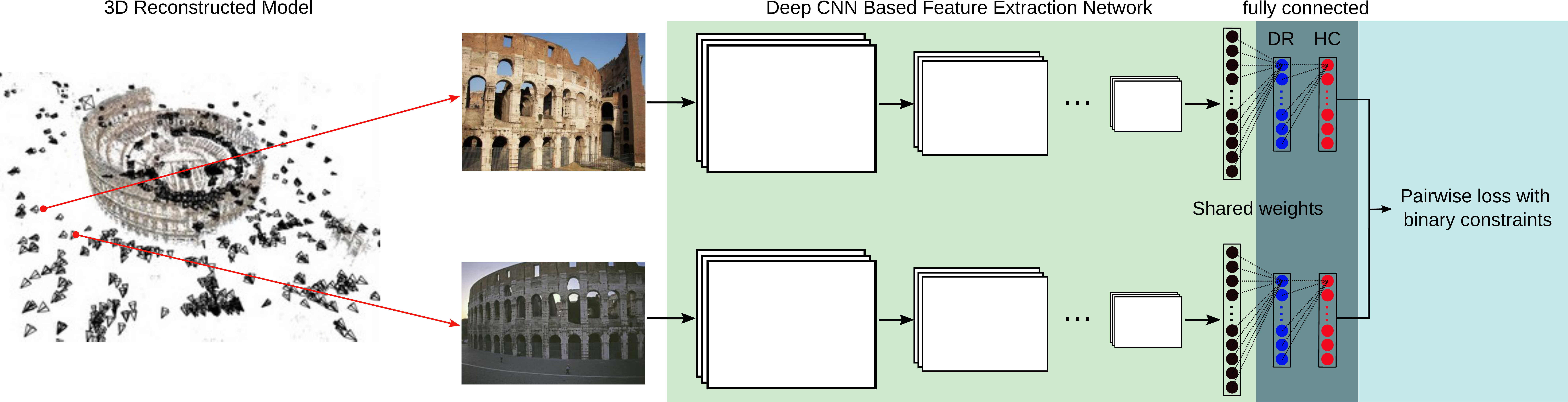}
\caption[]{\small The overview of the proposed deep hashing. Training data is created  automatically by exploiting 3D reconstructed models and their associated images. The network architecture comprises of four components: (i) convolutional layers which is followed by a MAC layer (black layer) for extracting image representations; (ii) a fully connected layer for Dimensional Reduction (blue layer); (iii) a Hash Code mapping layer (red layer); (iv) a novel pairwise loss function with binary constraints.}
\label{fig:model}
\end{figure*}
\label{sec:proposed}
Figure~\ref{fig:model} illustrates the proposed pipeline which trains a deep network for learning binary codes without the need of manually annotated data. In the following, we describe the proposed framework including: 
the network architecture, the automation of training data generation, 
the pairwise loss function, and the learning of the network parameters.

\input{tex/network_architecture.tex}
\input{tex/training_data.tex}
\input{tex/pairwise_loss.tex}
\input{tex/param_learning.tex}

%% file: tex/network_architecture.tex
\subsection{Network architecture}
The proposed network (Figure~\ref{fig:model}) comprises of four components: (i) a feature extraction component for extracting image representations; (ii) a fully connected layer for reducing dimension of the image features (Dimensionality Reduction -- DR layer). The number of  units of this layer equals to the code length required to represent each input image; (iii) a fully connected layer which maps the reduced real-valued features to binary values (Hash Code -- HC layer); 
(iv) a pairwise loss function which acts on the outputs of the HC layer. 
It is worth noting that the choice of the feature extraction component is flexible in our framework. It can be the standard convolutional neural network architectures, e.g. AlexNet \cite{DBLP:conf/nips/KrizhevskySH12} or VGG \cite{Simonyan14c}, in which the outputs of their last fully connected layer are used as inputs for the DR layer. Alternatively, the recent architecture which replaces the fully connected layers of VGG or AlexNet by a Maximum Activations of Convolutions (MAC) layer \cite{DBLP:journals/corr/ToliasSJ15} can also be used. In this case, the MAC features will be used as inputs for the DR layer.
For the image retrieval task focused in this paper, we adopt the VGG network with a MAC layer as we empirically find that using MAC layer gives better performance than fully connected layer.

Specifically, the MAC layer can be described as follows: Given an input image, the output of the last convolutional layer of VGG is a 3D tensor $W \times H \times K$, where $K$ is the number of output feature maps which equals to $512$, and $W \times H$ is spatial size of the last convolutional layer. Let $\mathcal{X}_k$ be $k^{th}$ feature map. The MAC image representation is constructed by
\begin{equation}
\u = \left[ u_1,\dots,u_k,\dots, u_K\right]^T, \textrm{where\ } u_k = \max_{x\in \mathcal{X}_k} x
\end{equation}

It is worth noting that the previous work \cite{DBLP:journals/corr/RadenovicTC16} also used a pairwise loss for fine-tuning deep network. However, our work significantly differs from \cite{DBLP:journals/corr/RadenovicTC16} at many aspects: while the target in~\cite{DBLP:journals/corr/RadenovicTC16} is to learn real-valued representations (i.e., 512-D real-valued features), our work aims to learn compact binary codes. To this end, we have two additional layers, i.e., dimensionality reduction and hash code, which are trained end-to-end together with other components. 
More important, our pairwise loss (Section~\ref{subsec:loss}) has binary constraints, unlike constraint-free loss as \cite{DBLP:journals/corr/RadenovicTC16}. The new layers and binary constrained loss function are crucial for hashing, i.e., it ensures the model to produce compact binary codes. Furthermore, we propose a learning scheme to cope with the binary constrained loss function (Section~\ref{subsec:opt}). This differs from \cite{DBLP:journals/corr/RadenovicTC16} which simply uses standard back-propagation to train the network with the constraint-free loss. 

%% file: tex/training_data.tex
\subsection{Training data}
\label{subsec:data}
The training input for our network is pairs of matching and non-matching images.  
One way to achieve training pairs is to access to semantic (class) labelled images so that we could train a hash function which returns similar hash codes for images with the same label, or vice versa. Unfortunately, such labelled dataset is not always available, especially for some problems which are not directly based on classification such as image retrieval. 
To overcome this challenge, we automatically create a training set of matching and non-matching image pairs by exploiting 3D reconstructed models and their associated images.  
In particular, we make use of the 3D models given by \cite{DBLP:journals/corr/RadenovicTC16}, in which there are $713$ 3D models reconstructed from  
images downloaded from Flickr. 
Most of reconstructed models are popular landmarks and cities. The authors released 3D models and 30K images which were used to build models. $5,974$ and $1,691$ images 
 are selected as training queries and validation queries, respectively.

In order to mining matching pairs, we follow the procedure used in \cite{DBLP:journals/corr/RadenovicTC16}. Given a training query image and its 3D model membership, we select images that belong to the same 3D model and co-observe enough 3D points. Among these, one image is randomly sampled and used as the matching image for the query. The set of matching pairs is kept during the training. 
In order to mining non-matching pairs, different from \cite{DBLP:journals/corr/RadenovicTC16}, we perform two stages of generation. The offline stage generates pairs for training the network at first iterations. After a certain iterations, we use the current network to perform the online generation (i.e., regenerating non-matching pairs) and use new pairs to continue the training. 
In particular, the offline generation is performed as follows: Given a training query image, we select top $k$ ``nearest'' images from 3D models different than the model containing the query. The distances between images are computed by using features extracted from the pretrained network \cite{DBLP:journals/corr/RadenovicTC16}. 
 Among these $k$ non-matching images, we randomly sample $m$ images (with at most one image per model) as the non-matching ones for the query. After a certain of iterations, we perform the online non-matching pair generation. The online generation is similar to the offline ones, except that the distance between images is computed by using the binary codes generated by the hash code layer of the current network. The values of $k$ and $m$ are empirically set to $70$ and $6$ in our experiments. 
Note that our non-matching mining strategy is different from~\cite{DBLP:journals/corr/RadenovicTC16} at two main aspects. Firstly, in~\cite{DBLP:journals/corr/RadenovicTC16}, the authors selected top $m$  nearest ones (i.e., hardest negative) from $k$ non-matching images. We empirically found that randomly selecting hard negative images gives better retrieval results than selecting the hardest ones. This is consistent with the observation in \cite{DBLP:conf/cvpr/SchroffKP15,DBLP:journals/corr/KumarHC0D17}, i.e., using the hardest negative samples can in practice lead to bad local minima in training. Secondly, because our target is to learn binary codes, hence in the online stage, we mine negative images using binary codes produced by the HC layer, rather than the features produced by the MAC layer as in~\cite{DBLP:journals/corr/RadenovicTC16}.

%% file: tex/pairwise_loss.tex
\subsection{Pairwise binary constrained loss}
\label{subsec:loss}

Given the training image pairs, we aim to train the network which not only produces binary codes but also ensures the discrimination of the codes, i.e., 
 matching images should likely have similar binary codes, or vice versa.
As the Hamming distance between two strings of binary codes is one-to-one corresponding to their Euclidean distance, we propose to minimize the following binary constrained loss function which acts on the pairs
\begin{equation}
\small
\min_{\W} \mathcal{L}(i,j) = y_{ij} \norm{\f_i - \f_j}_2 + \left(1-y_{ij}\right) \max\left( 0, c -\norm{\f_i - \f_j}_2\right)  \label{eq:loss_0}
\end{equation}
\vspace{-1em}
\begin{equation}
\textrm{s.t. }\f_i, \f_j \in \{ -1,1 \}^L \label{eq:binary}
\end{equation}
where $\W$ is the network parameters;
$\f_i$  and $\f_j$ are outputs of the Hash Code layer for input
 images $i$ and $j$, respectively; the label $y_{ij}\in \{0,1\}$ indicates that the image pair $i$, $j$ is matching ($y_{ij}=1$) or non-matching ($y_{ij}=0$); $L$ is the code length; $c$ is a constant.

Technically, the loss function will encourage matching pairs to have similar hash codes, and non-matching pairs to have different hash codes. When a non-matching pair has a large enough distance, i.e. $\ge c$, it is not to be taken into account in the loss.  
The constraint (\ref{eq:binary}) is to ensure the network outputs are binary. Optimizing the loss function (\ref{eq:loss_0}) with the binary constraint (\ref{eq:binary}) using stochastic gradient method is difficult since the constraints are not differentiable. To  overcome this challenge, we utilize the idea of the penalty method \cite{Nocedal06}. This leads to a formulation which avoids solving the exact binary constraint but instead minimizes the binary quantization loss. 
This makes sense because when the binary quantization loss approaches zero, the binary constraints are approximately satisfied.
Specifically, we introduce new auxiliary binary variables $\B=\{\bb_i\}_{i=1}^{N} \in \{-1,1\}^{L\times N}$ where $N$ is number of current training images, and
minimize the following loss function 

\footnotesize{
\begin{align}
\min_{\W,\B}\mathcal{L}(i,j) =& y_{ij} \norm{\f_i - \f_j}^2 + \left(1-y_{ij}\right) \left(\max\left( 0, c -\norm{\f_i - \f_j}\right)\right)^2 \nonumber \\
&+ \alpha \left( \norm{\f_i-\bb_i}^2 + \norm{\f_j - \bb_j}^2\right)\label{eq:loss_1}
\end{align}

\begin{equation}
\textrm{s.t. }\bb_i, \bb_j \in \{ -1,1 \}^L \label{eq:binary_1}
\end{equation}
}
\normalsize
\\
where $\alpha$ is a weighting parameter. The third term of (\ref{eq:loss_1}) forces the output of the Hash Code layer (i.e., $\f_i, \f_j$) as close to binary values as possible, i.e., it minimizes the binary quantization loss. 
Although the new loss function still contains constraints, the variables $\B$ and $\W$ are decoupled. This allows us to apply alternating optimization over these variables. We will show shortly that $\W$ now can be optimized using the stochastic gradient decent method and $\B$ has a closed-form solution. 

%% file: tex/param_learning.tex
\subsection{Parameter learning}
\label{subsec:opt} 
\begin{algorithm}[!t]

	\small
	\caption{Parameter learning}
	\begin{algorithmic}[1] 
		\Require 
			\Statex Reconstructed 3D models and their images; $L$: code length; $K$: number of online non-matching pairs generation; $T$: number of iterations for training network, given a fixed set of matching / non-matching pairs.
		\Ensure 
			\Statex 
			Set of network parameters $\{\W\}$. 
			\Statex 
			\State Offline generation of matching and non-matching pairs using the pretrained MAC network \cite{DBLP:journals/corr/RadenovicTC16}.
			\State Initialize the network $\W_{(0)}$
			\For{$k=1 \to K$}
			\For{$t = 1 \to T$}
				\State Fix $\W_{(t-1)}$, compute $\B_{(t)} = sgn(\F_{(t-1)})$
				\State Fix $\B_{(t)}$, optimize $\W_{(t)}$ (using $\W_{(t-1)}$ as initialization) using back-propagation. Save $\W_{(t)}$.
			\EndFor
			\State Regenerate non-matching pairs using $\W_{(T)}$
			\State Reinitialize $\W_{(0)} = \W_{(T)}$
			\EndFor
    \end{algorithmic}
    \label{alg1}
\end{algorithm}
In order to minimize the loss function \eqref{eq:loss_1} under the constraint \eqref{eq:binary_1}, we propose an alternating optimization approach, i.e., we learn each variable (network parameter $\W$ or $\B$) at a time while holding the other fixed. 
\begin{itemize}

\item Fix $\W$, solve $\B$: Let $\F = \{\f_i\}_{i=1}^N$, if $\W$ is fixed, 
the optimal solution for $\B$ is $sgn(\F)$.

\item Fix $\B$, solve $\W$: When $\B$ is fixed, the binary constraint \eqref{eq:binary_1} 
 can be ignored, thus, the network parameters $\W$ can be optimized by minimizing the loss  \eqref{eq:loss_1} using the standard back-propagation. 
A number of epochs is run until $\W$ converges to local minima before switching to  $\B$.
\end{itemize}
 The whole learning process is summarized in the Algorithm \ref{alg1}. In the Algorithm, $\W_{(t)}$, $\F_{(t)}$, $\B_{(t)}$ are values of $\W, \F, \B$ at $t^{th}$ iteration. 
We implement the proposed approach using the MatConvNet toolbox \cite{DBLP:journals/corr/VedaldiL14}. At the begining (line $2$ in the Algorithm), we initialize the feature extraction component with the pretrained MAC network \cite{DBLP:journals/corr/RadenovicTC16} in which its loss layer is removed. The DR layer is initialized by using PCA weights on the training dataset with pretrained MAC features.  

Inside the second \textit{for loop}, we fix the training pairs and alternative solving $\B$ and $\W$. When fixing $\B_{(t)}$ and learning $\W_{(t)}$ (line $6$ in the Algorithm), we train the network with a fix number of epochs $np$.  
 The values of $K$, $T$ and $np$ 
 are set to $4$, $5$ and $10$, respectively. The values of $c$ and $\alpha$ in (\ref{eq:loss_1}) are set to $\frac{L}{2}$ and $1$, respectively. 
 The batch size for learning $\W_{(t)}$ is $28$ pairs, i.e., 4 query images; each provides 1 matching pair and 6 non-matching pairs). Inside an iteration $k$, for each query, we generate $m=6$ non-matching pairs (line $8$ in the Algorithm). The best network is selected based on mean Average Precision (mAP) on the validation set.

%% file: tex/experiment.tex
\section{Experiments}
\label{sec:exp}
To evaluate the proposed method, which is dubbed as \textit{P2B (Pixels to Binary codes)},  we conduct extensive image retrieval experiments on standard image retrieval benchmarks. 

\begin{figure*}[!t]
\vspace{-0.4cm}
	\centering
	\begin{subfigure}[b]{0.31\textwidth}
		\includegraphics[width=\textwidth]{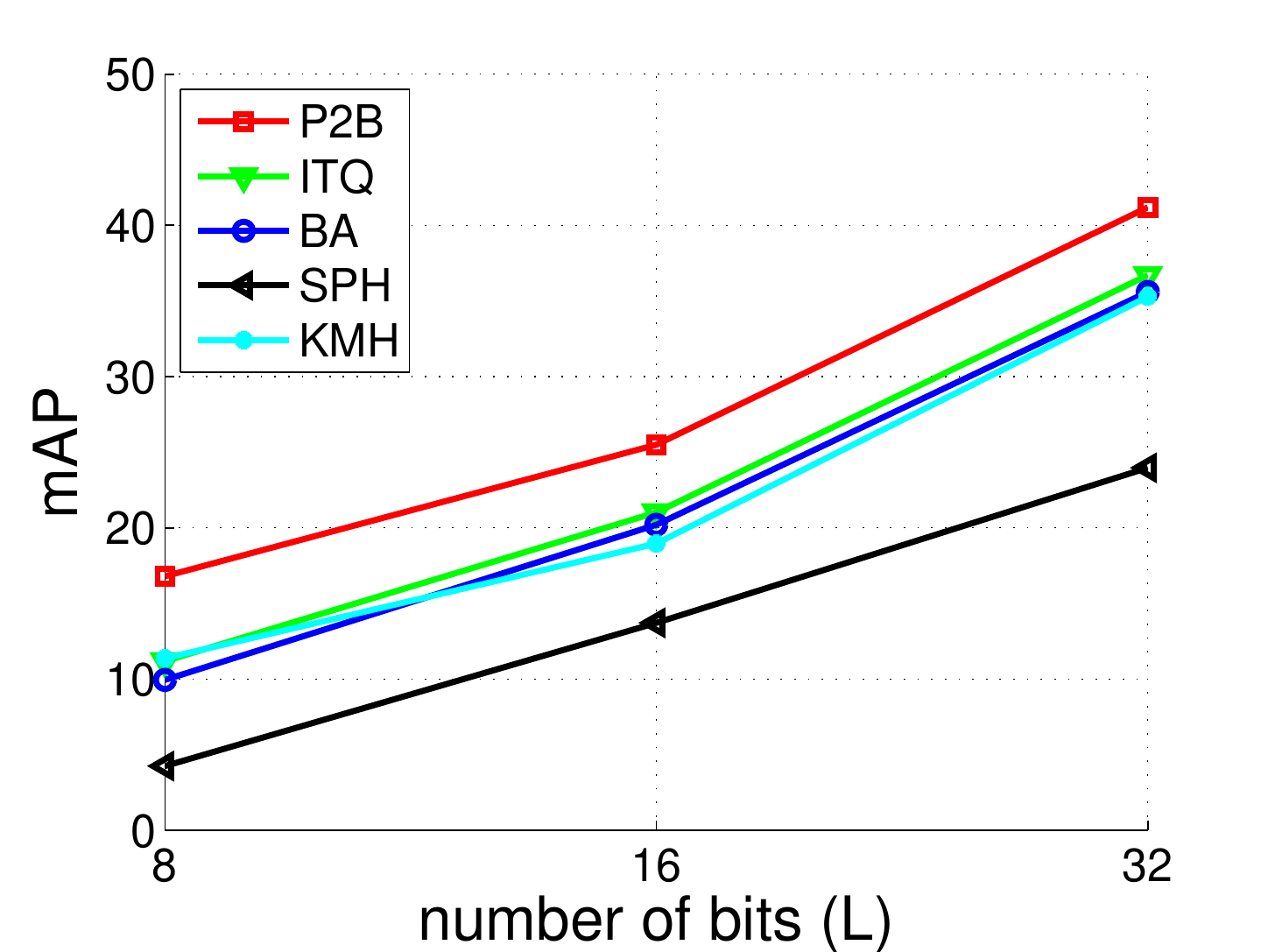}
		\caption{Oxford5k}
		\label{fig:ox5k_mAP}
	\end{subfigure}
	\begin{subfigure}[b]{0.31\textwidth}
		\includegraphics[width=\textwidth]{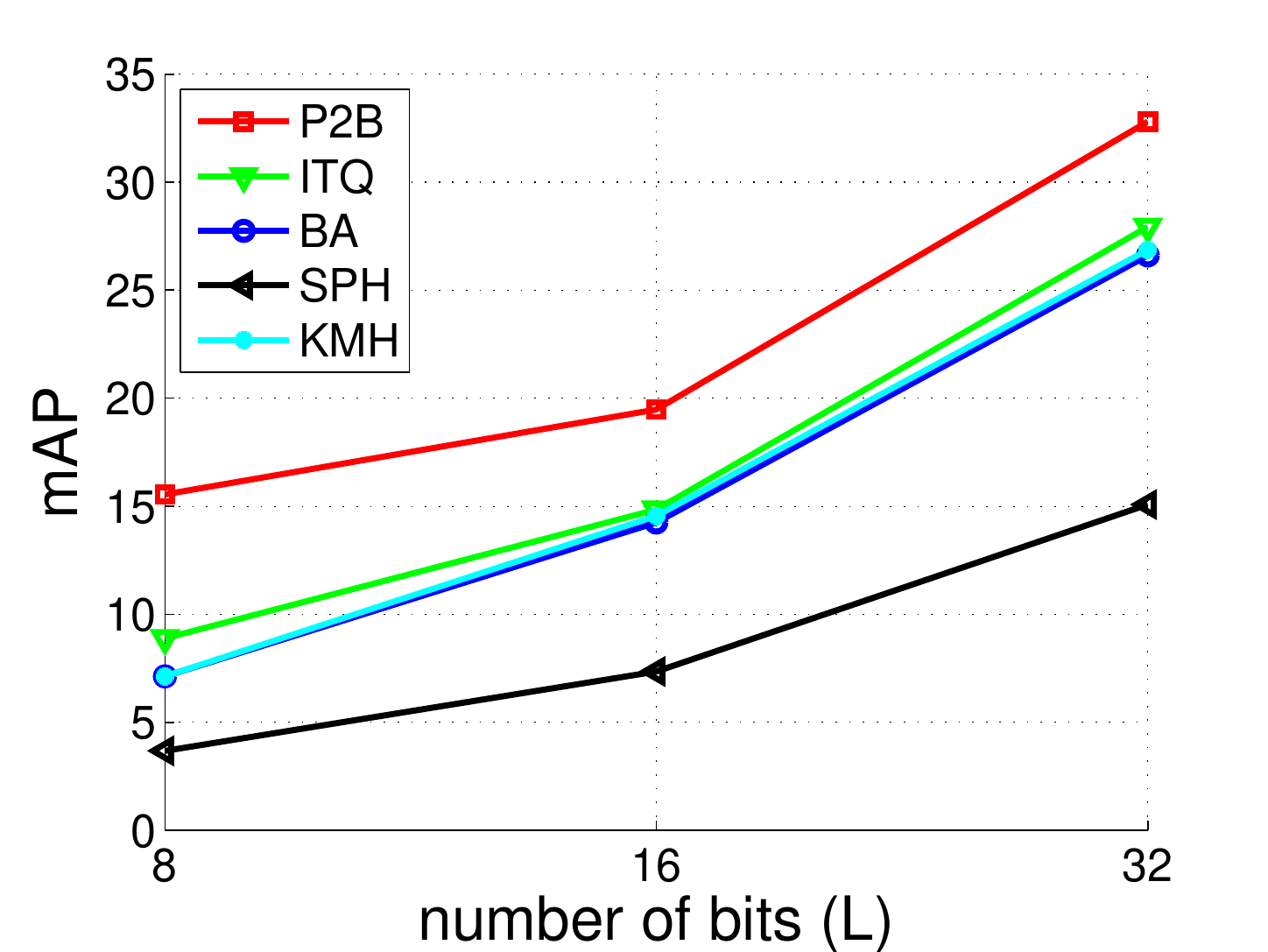}
		\caption{Oxford105k}
		\label{fig:oxford105k_mAP}
	\end{subfigure}
	\begin{subfigure}[b]{0.31\textwidth}
		\includegraphics[width=\textwidth]{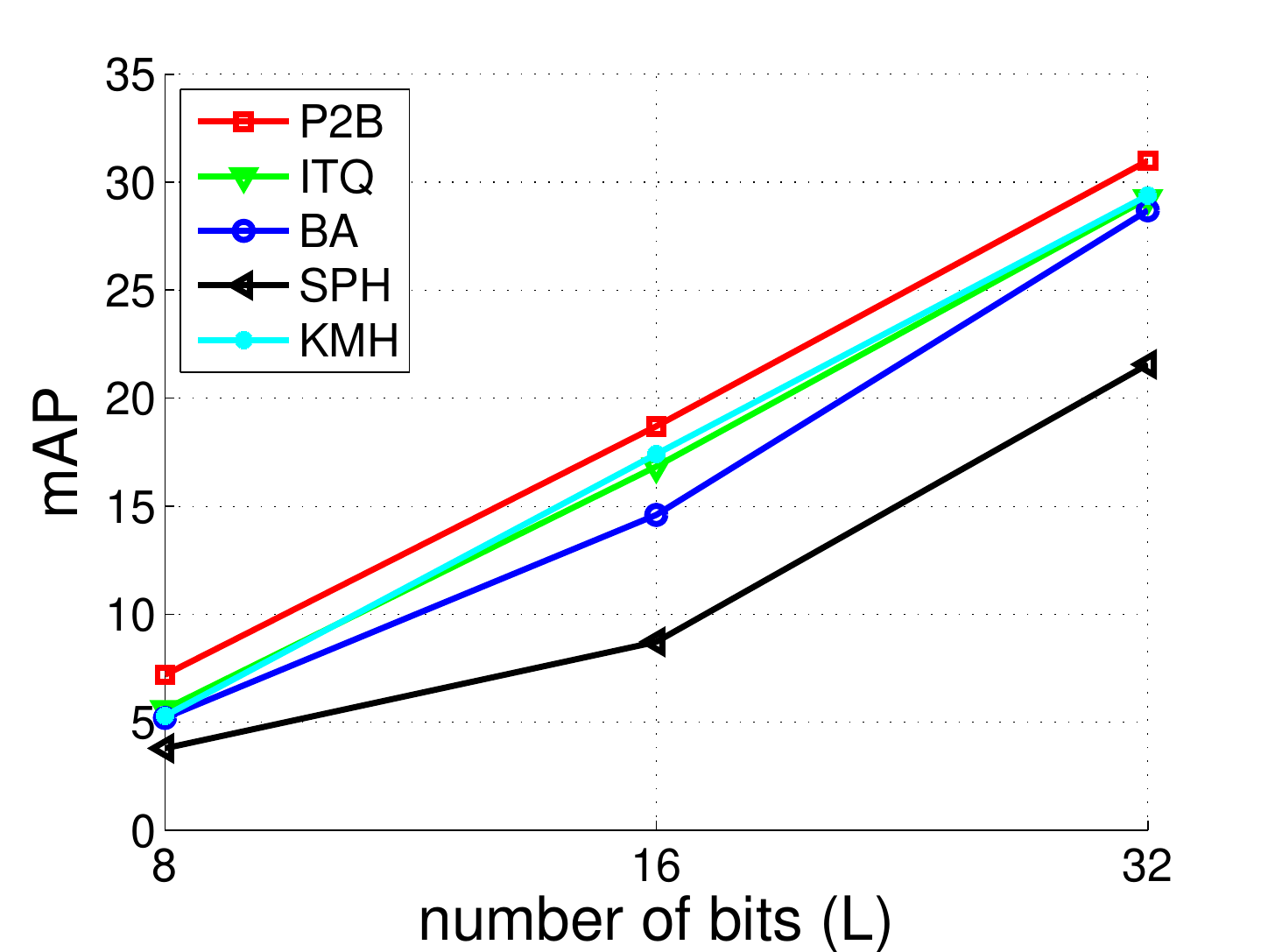}
		\caption{Holidays}
		\label{fig:holidays_mAP}
	\end{subfigure}
	\caption[]{mAP comparison between the proposed P2B and the traditional unsupervised hashing methods on Oxford5k, Oxford105k, and Holidays datasets. 
	}
	\label{fig:p2b-soa}
	\vspace{-0.4cm}
\end{figure*}
\subsection{Dataset and baselines}
\paragraph{Dataset}
We conduct experiments on Holidays~\cite{herve_ijcv2010}, Oxford5k~\cite{Philbin07-cvpr-2007} and Oxford105k~\cite{Philbin07-cvpr-2007} datasets which are widely used in evaluating image retrieval systems~\cite{DBLP:conf/cvpr/JegouZ14,DBLP:conf/cvpr/ArandjelovicZ13,herve_cvpr2010}. 

\textbf{Holidays} The Holidays dataset  
consists of 1,491 images of different locations and objects, 500 of them being used as queries. Most of images in dataset are natural scenes. Follow the standard protocol \cite{DBLP:conf/cvpr/JegouZ14,DBLP:conf/cvpr/ArandjelovicZ13}, when evaluating, we remove the query from the ranked list.

\textbf{Oxford5k} The Oxford5k dataset 
consists of 5,063 images of buildings and 55 query images corresponding to 11 distinct buildings in Oxford. Follow the standard protocol \cite{DBLP:conf/cvpr/JegouZ14,DBLP:conf/cvpr/ArandjelovicZ13}, we crop the bounding boxes of the region of interest and use them as the queries.

\textbf{Oxford105k} In order to evaluate the proposed method at larger scale, we merge Oxford5k dataset with 100k distracted images downloaded from Flickr~\cite{Philbin07-cvpr-2007}, forming the Oxford105k dataset.

The ground truth of queries have been provided with the datasets.  
Following previous works~\cite{BA_CVPR15,deepbit2018,DBLP:conf/aaai/SongHGXS18,Kevin_TPAMI17}, we evaluate the performance of methods at different code lengths, i.e. $8$, $16$, $32$, $256$, $512$ bits. The retrieval accuracy is measured by mean Average Precision (mAP).
\paragraph{Baselines}
We compare the proposed P2B against state-of-the-art unsupervised hashing methods, including both traditional methods: Iterative
Quantization (ITQ)~\cite{DBLP:conf/cvpr/GongL11}, Binary Autoencoder
(BA)~\cite{BA_CVPR15}, Spherical Hashing (SPH)~\cite{CVPR12:SphericalHashing},
K-means Hashing (KMH)~\cite{DBLP:conf/cvpr/HeWS13}, and recent deep-based methods: DeepBit~\cite{deepbit2018}, Deep Region Hashing (DRH)~\cite{DBLP:conf/aaai/SongHGXS18}. 

We also compare P2B against supervised hashing methods, including traditional methods: Kernel-based Supervised Hashing (KSH)~\cite{CVPR12:Hashing}, Binary Reconstructive Embedding (BRE)~\cite{Kulis_learningto}, ITQ-CCA~\cite{DBLP:conf/cvpr/GongL11}, and recent deep-based methods: Supervised Semantics-preserving Deep Hashing (SSDH)~\cite{Kevin_TPAMI17}, Hierachical Deep Hashing (HDH)~\cite{Song2017}. Note that, to the best of our knowledge, very few supervised hashing methods~\cite{Kevin_TPAMI17,Song2017} have evaluated on standard image retrieval datasets such as Holidays, Oxford5K, and Oxford105K. It may be because there are no available large scale labelled training data for those datasets.

Furthermore, we also compare P2B with the recent state-of-the-art real-valued image representations for the image retrieval problem. They are regional maximum activation of convolution (R-MAC)~\cite{DBLP:journals/corr/ToliasSJ15}, triangulation embedding (T-emb)~\cite{DBLP:conf/cvpr/JegouZ14}, function approximation-based embedding (F-FAemb)~\cite{f-faemb}, off-the-shelf CNN (OS-CNN)~\cite{DBLP:conf/cvpr/RazavianASC14}, faster-RCNN~\cite{DBLP:conf/cvpr/SalvadorNMS16}, neural codes~\cite{babenko2014neural}, sum pooling of convolutional feature (SPoC)~\cite{DBLP:conf/iccv/BabenkoL15}, cross-dimensional weighting (CDW)~\cite{DBLP:conf/eccv/KalantidisMO16},  unsupervised fine-tuning CNN (UF-CNN)~\cite{DBLP:journals/corr/RadenovicTC16},
regional attention based deep feature (RADF)~\cite{retrieval:BMVC:2018}, CNN-based VLAD (NetVLAD)~\cite{netvlad}, the end-to-end CNN (E2E-CNN)~\cite{DBLP:journals/ijcv/GordoARL17}.

\subsection{Comparison with unsupervised hashing methods}
\subsubsection{Comparison with traditional unsupervised hashing methods}
All compared traditional unsupervised methods require image features as inputs, instead of raw images. In order to make a fair comparison, we use the pretrained network~\cite{DBLP:journals/corr/RadenovicTC16} to extract the MAC features of 30K images which are used to reconstruct 3D models and use them as training inputs for traditional unsupervised hashing methods.

Figure~\ref{fig:p2b-soa} shows the comparative retrieval results between methods in term of mAP. On the Oxford5k and Oxford105k datasets, the results show that the proposed P2B significantly outperforms other methods, i.e., P2B outperforms the most competitive ITQ~\cite{DBLP:conf/cvpr/GongL11} $\ge 4.5\%$ mAP at all code lengths. The improvements are clearer at the lower code lengths, i.e., at $L=8$, P2B outperforms ITQ $5.6\%$ and $6.7\%$ mAP on Oxford5k and Oxford105k, respectively. 

On the Holidays dataset, the proposed P2B 
 also outperforms ITQ \cite{DBLP:conf/cvpr/GongL11} and other methods, i.e., P2B outperforms ITQ around 1.5\% to 2\% mAP at different code lengths. The results from Figure~\ref{fig:p2b-soa} also show that the improvements of P2B over other methods are clearer on the Oxford5k and Oxford105k datasets than on the Holidays dataset. The possible reason is that the Oxford5k building dataset may share similar visual characteristics (e.g., man-made architecture) with the training images which mostly contain landmarks and buildings. These datasets have different characteristics with the Holidays dataset which mostly contains images of natural scenes. 
 
\subsubsection{Comparison with unsupervised deep hashing methods}
Here we compare the proposed P2B with the DeepBit \cite{deepbit2018} and Deep Region Hashing (DRH)~\cite{DBLP:conf/aaai/SongHGXS18}, which are the state-of-the-art  deep learning-based unsupervised hashing methods. 
To make a fair comparison to DRH, we compare to its results at comparable code lengths (e.g., $L=256,512$) without query expansion\footnote{In DRH~\cite{DBLP:conf/aaai/SongHGXS18} the authors achieved best mAP at $85.1\%$ with multi-stage searching and query expansion at code length $L=4096$.}. Note that, in DRH the authors used a part of Oxford5k dataset to train their hashing layer which is a non-standard training setting when evaluating on Oxford5k~\cite{DBLP:conf/cvpr/JegouZ14,DBLP:conf/cvpr/ArandjelovicZ13}. In DeepBit~\cite{deepbit2018}, the authors trained their model using the semi-manually labelled landmark dataset~\cite{babenko2014neural} which is expected to have less noise level than our automatically created training dataset. 
\begin{table}[!t]
	\begin{center}
		\begin{tabular}{|c|c c|}
			\hline
			L				 &256	 &512   \\\hline
			DeepBit~\cite{deepbit2018}  &60.30 &62.70 \\
			DRH~\cite{DBLP:conf/aaai/SongHGXS18}  &58.30 &66.80 \\
			P2B                  &69.20 &74.84 \\\hline
		\end{tabular}
	\end{center}
	\caption{mAP results of P2B, DeepBit~\cite{deepbit2018}, DRH~\cite{DBLP:conf/aaai/SongHGXS18} on Oxford5k. The results of the compared methods are cited from the corresponding papers.}
	\label{tab:compare_deepbit}
	\vspace{-1em}
\end{table}
The results from the Table~\ref{tab:compare_deepbit} show that, at the comparable code lengths, the proposed P2B significantly outperforms the compared methods. P2B outperforms DeepBit $12.1\%$ and outperforms DRH $8\%$ mAP at $L=512$, even when DRH is trained on images from the Oxford5k dataset and DeepBit is trained on the semi-manually labelled landmark dataset.

\input{tex/exp_sup2.tex}

\subsection{Comparison with real-valued image representations}

\begin{table}[!t]
	\begin{center}
		\begin{tabular}{|c|c|c|}
			\hline
			Methods		&D		 &mAP   \\\hline
R-MAC~\cite{DBLP:journals/corr/ToliasSJ15} &512 (float) &66.9 \\
T-emb~\cite{DBLP:conf/cvpr/JegouZ14}  &1024 (float) &56.2 \\
F-FAemb~\cite{f-faemb}				&1024 (float) &58.2 \\
OS-CNN~\cite{DBLP:conf/cvpr/RazavianASC14} &4096 (float)&68.0 \\
Faster-RCNN~\cite{DBLP:conf/cvpr/SalvadorNMS16} &4096 (float)&67.8 \\
Neural codes~\cite{babenko2014neural} &256 (float)&55.7 \\
SPoC~\cite{DBLP:conf/iccv/BabenkoL15} &256 (float) &58.9 \\
CDW~\cite{DBLP:conf/eccv/KalantidisMO16} &256 (float) &65.4 \\
NetVLAD~\cite{netvlad}  &256 (float) &63.5 \\
UF-CNN~\cite{DBLP:journals/corr/RadenovicTC16} &512 (float) &79.7 \\
RADF~\cite{retrieval:BMVC:2018}  &2048 (float)&76.8 \\
E2E-CNN~\cite{DBLP:journals/ijcv/GordoARL17} &2048 (float) &86.1 \\
P2B						&512 (bit) &74.8 \\\hline
		\end{tabular}
	\end{center}
	\caption{Comparative mAP between P2B and the state-of-the-art real-valued representations on Oxford5k dataset. The second column is dimensionality of the presentations. Note that the compared methods uses real-valued ($float$) representations, while P2B uses binary ($bit$) presentations.}
	\label{tab:compare_compact}
\end{table}
In this section we compare the proposed P2B to the state-of-the-art real-valued image representations. Table~\ref{tab:compare_compact} presents the comparative mAP between methods on Oxford5k dataset. 
The results show that, even using binary representation, P2B outperforms most compared methods that use the real-valued representations. The state-of-the-art mAP is achieved by the real-valued high dimensional representation E2E-CNN~\cite{DBLP:journals/ijcv/GordoARL17}. In particular, E2E-CNN outperforms P2B $11.3\%$ mAP. 
Although the proposed P2B does not provide the state-of-the-art retrieval accuracy, it is worth mentioning that in term of memory, P2B uses only $64$ bytes to represent an image, which is $128$ times less than the one of E2E-CNN. Furthermore, the binary representation of P2B also allows the fast distance calculation, i.e., Hamming distance. This makes P2B more suitable than E2E-CNN for the large scale retrieval problem. 

%% file: tex/exp_sup2.tex
\subsection{Comparison with supervised hashing methods}
\label{compare_supervised}

\begin{figure*}[!h]
\centering
\begin{subfigure}[b]{0.33\textwidth}
\includegraphics[width=\textwidth]{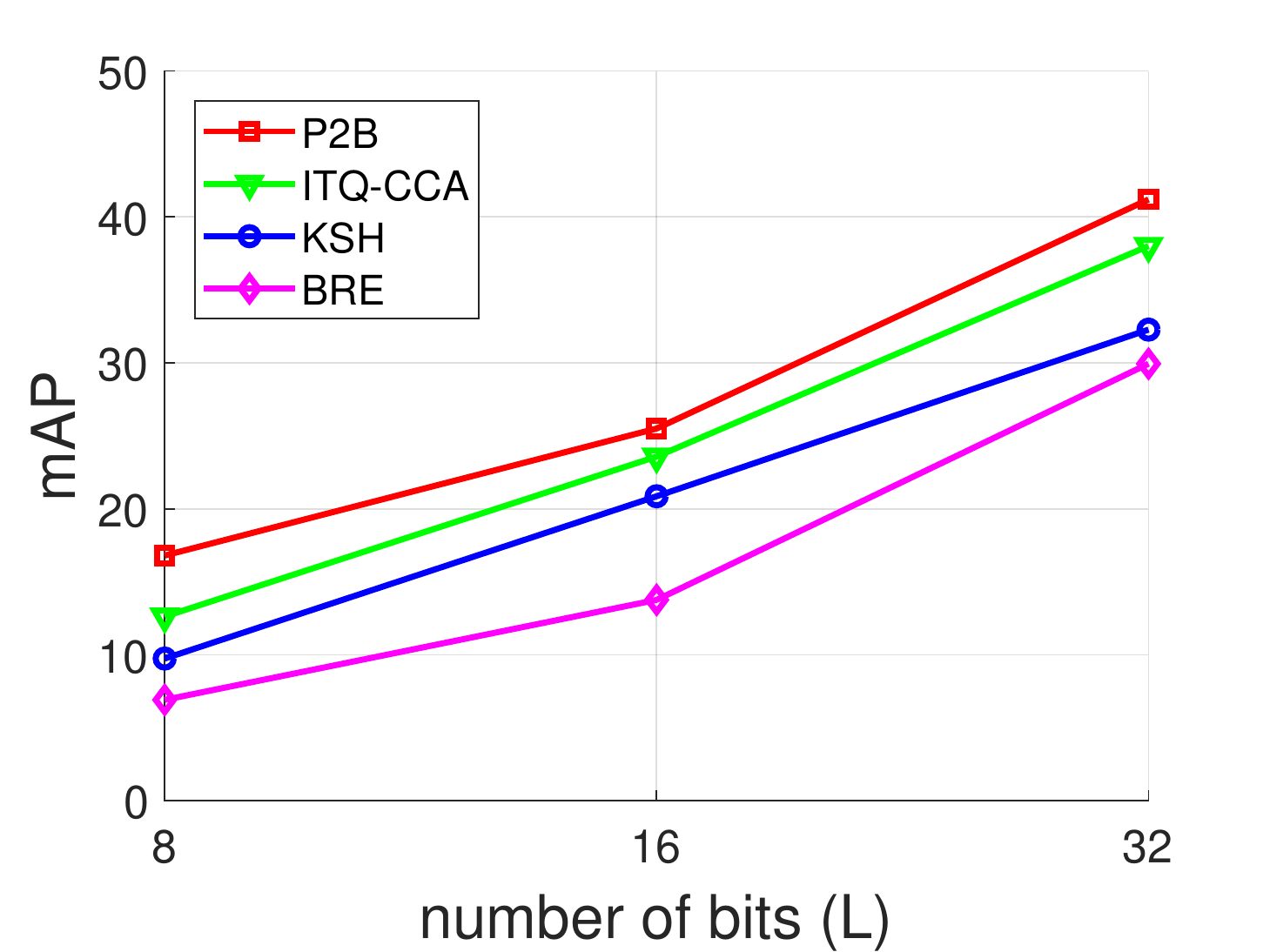}
\caption{Oxford5k}
\label{fig:sup_ox5k_mAP}
\end{subfigure}
\begin{subfigure}[b]{0.33\textwidth}
\includegraphics[width=\textwidth]{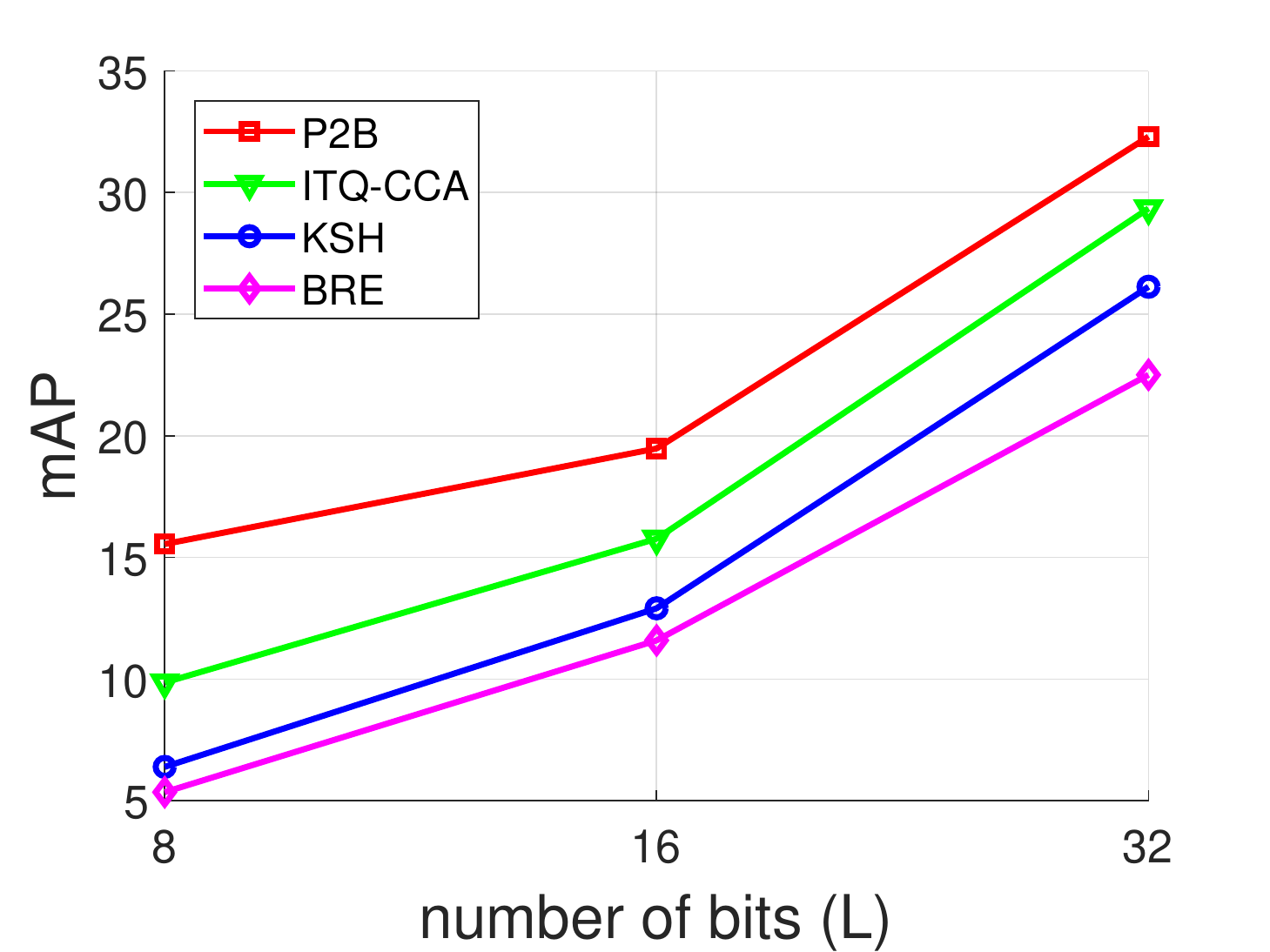}
\caption{Oxford105k}
\label{fig:sup_oxford105k_mAP}
\end{subfigure}
\begin{subfigure}[b]{0.33\textwidth}
\includegraphics[width=\textwidth]{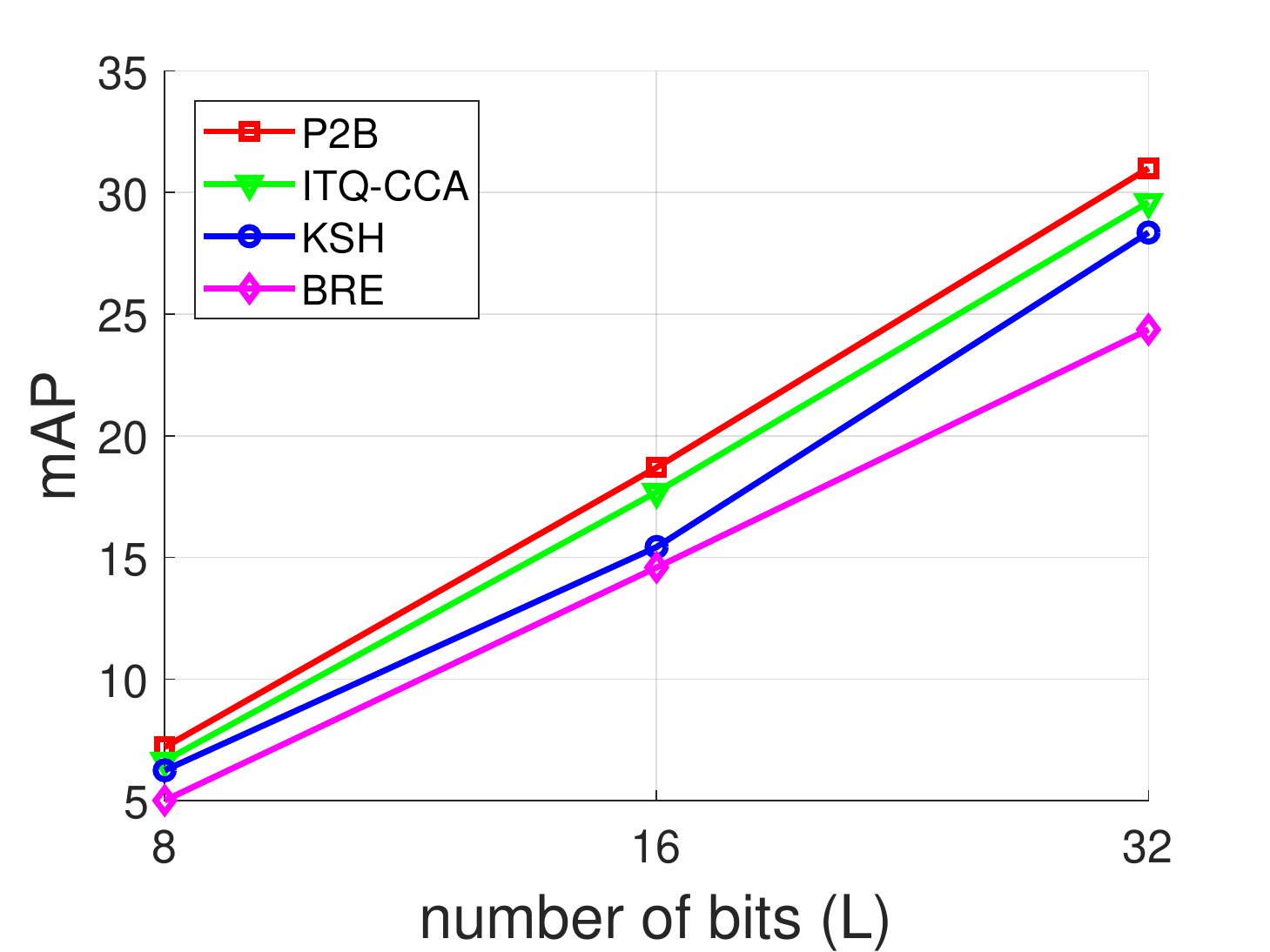}
\caption{Holidays}
\label{fig:sup_holidays_mAP}
\end{subfigure}
\caption[]{mAP comparison between the proposed P2B and the traditional supervised hashing methods on Oxford5k, Oxford105k, and Holidays datasets. 
}
\label{fig:p2b-soa-sup}
\end{figure*}

\subsubsection{Comparison with traditional supervised hashing methods}
The supervised hashing methods, such as KSH, BRE, CCA-ITQ, require the class label to perform the training. 
 To make a fair comparison, we use the images which are used to reconstruct the 3D models and carefully investigate different strategies to define the sample similarity when training  traditional supervised hashing methods.

As the number of images of each 3D model varies, i.e., the largest model contains 80 images, while the smallest model contains only 23 images, we first select top 100 biggest 3D models and then randomly select 60 images per model for training. 
Note that as KSH and BRE require the full similarity between every pair of samples when training, it is difficult for these methods to handle larger training data. 
Let the similarity matrix be $S$, we try two approaches to define the similarity for each image pair in the matrix $S$. 

In the first approach, we check every image pair in the matrix $S$ and use the same matching pair generation strategy in Section 3.2 for determining matching pairs. 
A pair $(i,j)$ is matched, i.e. $S(i,j)=1$, if both images belong to the same 3D model and co-observe enough 3D points, otherwise $S(i,j)=0$. By using this first approach, we found that the matrix $S$ is very sparse. In the second approach, images which are belonged to the same model are considered as matching, otherwise, non matching. In the other words, each 3D model is considered as a class.
We empirically found that for KSH and BRE, the similarity matrix constructed by the second approach gives better retrieval results than the first approach, e.g., for KSH on Oxford5k dataset, the mAP at $L=32$ is $19.11\%$ and $32.28\%$ for the first and the second approach, respectively. We found that although the first approach uses better strategy for creating matching pairs, the matrix $S$ is very sparse. Hence the non-matching pairs strongly dominate the matching pairs during training, leading to poor results. In the following, we consistently use the second approach, which considers each 3D model as a class, for training the  compared methods. 

Figure \ref{fig:p2b-soa-sup} presents the comparative retrieval results of P2B, KSH, BRE, CCA-ITQ. The results show that on the Oxford5k and Oxford105 datasets, P2B outperforms compared methods with fair margin at all code lengths. On the Holidays dataset, P2B and CCA-ITQ achieve comparable results and these methods outperform KSH and BRE. 

\subsubsection{Comparison with supervised deep hashing methods}

\begin{table}[!t]
	\begin{center}
		\begin{tabular}{|c|c|c|}
			\hline
			L				             &256    &512\\\hline
			SSDH \cite{Kevin_TPAMI17}    &-      &63.80    \\\hline
			HDH \cite{Song2017}          &69.70  &70.50    \\\hline
			P2B                          &69.20  &74.84    \\\hline
		\end{tabular}
	\end{center}
	\caption{Comparative mAP between the proposed P2B and recent supervised deep hashing methods on the Oxford5k dataset. The results of the compared methods are cited from the corresponding papers.}
	\label{tab:compare_ssdh}
\end{table}

There are very few deep supervised hashing methods that report results on image retrieval benchmarks such as Holidays and Oxford5k. In this section we compare the proposed P2B to the recent deep supervised hashing methods  SSDH~\cite{Kevin_TPAMI17}, hierarchical deep hashing (HDH)~\cite{Song2017} which have reported their results on Oxford5k dataset. 
Table~\ref{tab:compare_ssdh} presents the comparative mAP between methods. 
It is worth mentioning that in SSDH~\cite{Kevin_TPAMI17}, the authors fine-tuned their model on a semi-manually labelled landmark dataset~\cite{babenko2014neural}, while in HDH~\cite{Song2017}, the authors fine-tuned their model using a part of the Oxford5k dataset, which is a non-standard training setting~\cite{DBLP:conf/cvpr/JegouZ14,DBLP:conf/cvpr/ArandjelovicZ13}. Diffrent from those works, our method uses automatically created training data from unlabeled images. The results in Table~\ref{tab:compare_ssdh} show that P2B outperforms SSDH a large margin, i.e., $11\%$ mAP at $L=512$. Compare to HDH, P2B and HDH achieve comparable mAP at $L=256$, while at higher code length, i.e. $L=512$, P2B improves $4.3\%$ mAP over HDH. It is also worth mentioning that our automatically created training data has higher noise level than the semi-manually labelled dataset used in SSDH. Specifically, we fine-tuned SSDH with the images used to reconstruct 3D models, i.e., we select top 100 biggest 3D models and consider each 3D model as a class. The mAP of the fine-tuned SSDH using these images at $L=512$ is $59.17\%$ which is lower than the reported $63.80\%$ in~\cite{Kevin_TPAMI17}. Inspite of the higher noise level in the training data, P2B achieves a mAP $74.84\%$ which is higher than the reported $63.80\%$ of SSDH. This confirms the effectiveness of the proposed framework.

%% file: tex/conclusion2.tex
\section{Conclusion}
\label{sec:concl}
In this paper, we propose a novel end-to-end deep hashing framework for directly learning compact binary codes from images without the need of manual annotation. We exploit the reconstructed 3D models and their associated images to automatically create the training data. We propose a novel pairwise binary constrained loss function which not only allows to leverage the discriminative information from training data but also encourages the output codes to be binary. We also propose an efficient alternating optimization to train the network under the constrained loss. 
The solid experimental results on image retrieval benchmark datasets show that the proposed method compares favorably with  the state of the art. 

%% file: 315.bbl
\begin{thebibliography}{10}\itemsep=-1pt

\bibitem{netvlad}
R.~Arandjelovic, P.~Gron{\'{a}}t, A.~Torii, T.~Pajdla, and J.~Sivic.
\newblock Netvlad: {CNN} architecture for weakly supervised place recognition.
\newblock {\em TPAMI}, pages 1437--1451, 2018.

\bibitem{DBLP:conf/cvpr/ArandjelovicZ13}
R.~Arandjelovic and A.~Zisserman.
\newblock All about {VLAD}.
\newblock In {\em CVPR}, 2013.

\bibitem{DBLP:conf/cvpr/AzizpourRSMC15}
H.~Azizpour, A.~S. Razavian, J.~Sullivan, A.~Maki, and S.~Carlsson.
\newblock From generic to specific deep representations for visual recognition.
\newblock In {\em CVPRW}, 2015.

\bibitem{DBLP:conf/iccv/BabenkoL15}
A.~Babenko and V.~S. Lempitsky.
\newblock Aggregating local deep features for image retrieval.
\newblock In {\em ICCV}, 2015.

\bibitem{babenko2014neural}
A.~Babenko, A.~Slesarev, A.~Chigorin, and V.~Lempitsky.
\newblock Neural codes for image retrieval.
\newblock In {\em ECCV}, 2014.

\bibitem{BA_CVPR15}
M.~A. Carreira-Perpinan and R.~Raziperchikolaei.
\newblock Hashing with binary autoencoders.
\newblock In {\em CVPR}, 2015.

\bibitem{f-faemb}
T.-T. Do and N.-M. Cheung.
\newblock Embedding based on function approximation for large scale image
  search.
\newblock {\em TPAMI}, pages 626--638, 2018.

\bibitem{UH-BDNN-Do2016-0}
T.-T. Do, A.-D. Doan, and N.-M. Cheung.
\newblock Learning to hash with binary deep neural network.
\newblock In {\em ECCV}, 2016.

\bibitem{DBLP:journals/tmm/ErcoliBB17}
S.~Ercoli, M.~Bertini, and A.~D. Bimbo.
\newblock Compact hash codes for efficient visual descriptors retrieval in
  large scale databases.
\newblock {\em TMM}, pages 2521--2532, 2017.

\bibitem{MAC2}
O.~C. Filip~Radenovic, Giorgos~Tolias.
\newblock Fine-tuning cnn image retrieval with no human annotation.
\newblock {\em TPAMI}, 2018.

\bibitem{lsh_vldb09}
A.~Gionis, P.~Indyk, and R.~Motwani.
\newblock Similarity search in high dimensions via hashing.
\newblock In {\em VLDB}, 1999.

\bibitem{DBLP:conf/cvpr/GongL11}
Y.~Gong and S.~Lazebnik.
\newblock Iterative quantization: {A} procrustean approach to learning binary
  codes.
\newblock In {\em CVPR}, 2011.

\bibitem{DBLP:journals/ijcv/GordoARL17}
A.~Gordo, J.~Almaz{\'{a}}n, J.~Revaud, and D.~Larlus.
\newblock End-to-end learning of deep visual representations for image
  retrieval.
\newblock {\em IJCV}, pages 237--254, 2017.

\bibitem{Grauman_review}
K.~Grauman and R.~Fergus.
\newblock Learning binary hash codes for large-scale image search.
\newblock {\em Machine Learning for Computer Vision}, 2013.

\bibitem{DBLP:conf/cvpr/HeWS13}
K.~He, F.~Wen, and J.~Sun.
\newblock K-means hashing: An affinity-preserving quantization method for
  learning binary compact codes.
\newblock In {\em CVPR}, 2013.

\bibitem{CVPR12:SphericalHashing}
J.-P. Heo, Y.~Lee, J.~He, S.-F. Chang, and S.-e. Yoon.
\newblock Spherical hashing.
\newblock In {\em CVPR}, 2012.

\bibitem{herve_ijcv2010}
H.~J\'egou, M.~Douze, and C.~Schmid.
\newblock Improving bag-of-features for large scale image search.
\newblock {\em IJCV}, pages 316--336, 2010.

\bibitem{herve_cvpr2010}
H.~J\'egou, M.~Douze, C.~Schmid, and P.~P\'erez.
\newblock Aggregating local descriptors into a compact image representation.
\newblock In {\em CVPR}, 2010.

\bibitem{DBLP:conf/cvpr/JegouZ14}
H.~J{\'{e}}gou and A.~Zisserman.
\newblock Triangulation embedding and democratic aggregation for image search.
\newblock In {\em CVPR}, 2014.

\bibitem{DBLP:conf/eccv/KalantidisMO16}
Y.~Kalantidis, C.~Mellina, and S.~Osindero.
\newblock Cross-dimensional weighting for aggregated deep convolutional
  features.
\newblock In {\em {ECCV} Workshops}, 2016.

\bibitem{retrieval:BMVC:2018}
J.~Kim and S.-E. Yoon.
\newblock Regional attention based deep feature for image retrieval.
\newblock In {\em British Machine Vision Conference (BMVC)}, 2018.

\bibitem{DBLP:conf/nips/KrizhevskySH12}
A.~Krizhevsky, I.~Sutskever, and G.~E. Hinton.
\newblock Imagenet classification with deep convolutional neural networks.
\newblock In {\em NIPS}, 2012.

\bibitem{Kulis_learningto}
B.~Kulis and T.~Darrell.
\newblock Learning to hash with binary reconstructive embeddings.
\newblock In {\em NIPS}, 2009.

\bibitem{KLSH_iccv09}
B.~Kulis and K.~Grauman.
\newblock Kernelized locality-sensitive hashing for scalable image search.
\newblock In {\em ICCV}, 2009.

\bibitem{DBLP:journals/pami/KulisJG09}
B.~Kulis, P.~Jain, and K.~Grauman.
\newblock Fast similarity search for learned metrics.
\newblock {\em TPAMI}, pages 2143--2157, 2009.

\bibitem{DBLP:journals/corr/KumarHC0D17}
B.~G.~V. Kumar, B.~Harwood, G.~Carneiro, I.~D. Reid, and T.~Drummond.
\newblock Smart mining for deep metric learning.
\newblock In {\em ICCV}, 2017.

\bibitem{DBLP:conf/cvpr/LaiPLY15}
H.~Lai, Y.~Pan, Y.~Liu, and S.~Yan.
\newblock Simultaneous feature learning and hash coding with deep neural
  networks.
\newblock In {\em CVPR}, 2015.

\bibitem{Lecun98gradient-basedlearning}
Y.~Lecun, L.~Bottou, Y.~Bengio, and P.~Haffner.
\newblock Gradient-based learning applied to document recognition.
\newblock {\em Proceedings of the IEEE}, pages 2278--2324, 1998.

\bibitem{CVPR2014Lin}
G.~Lin, C.~Shen, Q.~Shi, A.~{van den Hengel}, and D.~Suter.
\newblock Fast supervised hashing with decision trees for high-dimensional
  data.
\newblock In {\em CVPR}, 2014.

\bibitem{deepbit2016}
K.~Lin, J.~Lu, C.-S. Chen, and J.~Zhou.
\newblock Learning compact binary descriptors with unsupervised deep neural
  networks.
\newblock In {\em CVPR}, 2016.

\bibitem{deepbit2018}
K.~Lin, J.~Lu, C.-S. Chen, J.~Zhou, and M.-T. Sun.
\newblock Unsupervised deep learning of compact binary descriptors.
\newblock {\em TPAMI}, 2018.

\bibitem{DBLP:conf/cvpr/Liu0SC16}
H.~Liu, R.~Wang, S.~Shan, and X.~Chen.
\newblock Deep supervised hashing for fast image retrieval.
\newblock In {\em CVPR}, 2016.

\bibitem{CVPR12:Hashing}
W.~Liu, J.~Wang, R.~Ji, Y.-G. Jiang, and S.-F. Chang.
\newblock Supervised hashing with kernels.
\newblock In {\em CVPR}, 2012.

\bibitem{DeepHash_TIP17}
J.~Lu, V.~E. Liong, and J.~Zhou.
\newblock Deep hashing for scalable image search.
\newblock {\em TIP}, 2017.

\bibitem{Nocedal06}
J.~Nocedal and S.~J. Wright.
\newblock {\em Numerical Optimization}.
\newblock World Scientific, 2nd edition, 2006.

\bibitem{DBLP:conf/icml/NorouziF11}
M.~Norouzi and D.~J. Fleet.
\newblock Minimal loss hashing for compact binary codes.
\newblock In {\em ICML}, 2011.

\bibitem{Philbin07-cvpr-2007}
J.~Philbin, O.~Chum, M.~Isard, J.~Sivic, and A.~Zisserman.
\newblock Object retrieval with large vocabularies and fast spatial matching.
\newblock In {\em CVPR}, 2007.

\bibitem{DBLP:conf/cvpr/RadenovicSJFCM16}
F.~Radenovic, J.~L. Sch{\"{o}}nberger, D.~Ji, J.~Frahm, O.~Chum, and J.~Matas.
\newblock From dusk till dawn: Modeling in the dark.
\newblock In {\em CVPR}, 2016.

\bibitem{DBLP:journals/corr/RadenovicTC16}
F.~Radenovic, G.~Tolias, and O.~Chum.
\newblock {CNN} image retrieval learns from bow: Unsupervised fine-tuning with
  hard examples.
\newblock In {\em ECCV}, 2016.

\bibitem{KLSH_nips09}
M.~Raginsky and S.~Lazebnik.
\newblock Locality-sensitive binary codes from shift-invariant kernels.
\newblock In {\em NIPS}, 2009.

\bibitem{DBLP:conf/cvpr/RazavianASC14}
A.~S. Razavian, H.~Azizpour, J.~Sullivan, and S.~Carlsson.
\newblock {CNN} features off-the-shelf: An astounding baseline for recognition.
\newblock In {\em CVPRW}, 2014.

\bibitem{DBLP:conf/cvpr/SalvadorNMS16}
A.~Salvador, X.~{Gir{\'{o}}-i-Nieto}, F.~Marqu{\'{e}}s, and S.~Satoh.
\newblock Faster {R-CNN} features for instance search.
\newblock In {\em {CVPR} Workshops}, 2016.

\bibitem{DBLP:conf/cvpr/SchroffKP15}
F.~Schroff, D.~Kalenichenko, and J.~Philbin.
\newblock Facenet: {A} unified embedding for face recognition and clustering.
\newblock In {\em CVPR}, 2015.

\bibitem{Shen_2015_CVPR}
F.~Shen, C.~Shen, W.~Liu, and H.~Tao~Shen.
\newblock Supervised discrete hashing.
\newblock In {\em CVPR}, 2015.

\bibitem{SADH}
F.~Shen, Y.~Xu, L.~Liu, Y.~Yang, Z.~Huang, and H.~T. Shen.
\newblock Unsupervised deep hashing with similarity-adaptive and discrete
  optimization.
\newblock {\em TPAMI}, 2018.

\bibitem{Simonyan14c}
K.~Simonyan and A.~Zisserman.
\newblock Very deep convolutional networks for large-scale image recognition.
\newblock {\em CoRR}, 2014.

\bibitem{Song2017}
G.~Song and X.~Tan.
\newblock Hierarchical deep hashing for image retrieval.
\newblock {\em Frontiers of Computer Science, Springer}, pages 253--265, 2017.

\bibitem{DBLP:conf/aaai/SongHGXS18}
J.~Song, T.~He, L.~Gao, X.~Xu, and H.~T. Shen.
\newblock Deep region hashing for generic instance search from images.
\newblock In {\em AAAI}, 2018.

\bibitem{DBLP:journals/corr/ToliasSJ15}
G.~Tolias, R.~Sicre, and H.~J{\'{e}}gou.
\newblock Particular object retrieval with integral max-pooling of {CNN}
  activations.
\newblock In {\em ICLR}, 2016.

\bibitem{DBLP:journals/corr/VedaldiL14}
A.~Vedaldi and K.~Lenc.
\newblock Matconvnet - convolutional neural networks for {MATLAB}.
\newblock {\em CoRR}, 2014.

\bibitem{DBLP:journals/pami/WangKC12}
J.~Wang, S.~Kumar, and S.~Chang.
\newblock Semi-supervised hashing for large-scale search.
\newblock {\em TPAMI}, pages 2393--2406, 2012.

\bibitem{DBLP:journals/corr/WangLKC15}
J.~Wang, W.~Liu, S.~Kumar, and S.~Chang.
\newblock Learning to hash for indexing big data - {A} survey.
\newblock {\em Proceedings of the IEEE}, 2015.

\bibitem{DBLP:journals/corr/WangSSJ14_journal}
J.~Wang, T.~Zhang, J.~Song, N.~Sebe, and H.~T. Shen.
\newblock A survey on learning to hash.
\newblock {\em TPAMI}, 2017.

\bibitem{DBLP:conf/nips/WeissTF08}
Y.~Weiss, A.~Torralba, and R.~Fergus.
\newblock Spectral hashing.
\newblock In {\em NIPS}, 2008.

\bibitem{DBLP:conf/aaai/XiaPLLY14}
R.~Xia, Y.~Pan, H.~Lai, C.~Liu, and S.~Yan.
\newblock Supervised hashing for image retrieval via image representation
  learning.
\newblock In {\em AAAI}, 2014.

\bibitem{Kevin_TPAMI17}
H.-F. Yang, K.~Lin, and C.-S. Chen.
\newblock Supervised learning of semantics-preserving hash via deep
  convolutional neural networks.
\newblock {\em TPAMI}, pages 437--451, 2018.

\bibitem{DBLP:journals/tip/ZhangLZZZ15}
R.~Zhang, L.~Lin, R.~Zhang, W.~Zuo, and L.~Zhang.
\newblock Bit-scalable deep hashing with regularized similarity learning for
  image retrieval and person re-identification.
\newblock {\em IEEE TIP}, pages 4766--4779, 2015.

\bibitem{DBLP:conf/cvpr/ZhaoHWT15}
F.~Zhao, Y.~Huang, L.~Wang, and T.~Tan.
\newblock Deep semantic ranking based hashing for multi-label image retrieval.
\newblock In {\em CVPR}, 2015.

\bibitem{DBLP:conf/cvpr/ZhuangLSR16}
B.~Zhuang, G.~Lin, C.~Shen, and I.~D. Reid.
\newblock Fast training of triplet-based deep binary embedding networks.
\newblock In {\em CVPR}, 2016.

\end{thebibliography}
